\documentclass[11pt]{article}

\usepackage[final]{acl}

\usepackage{times}
\usepackage{latexsym}

\usepackage[T1]{fontenc}

\usepackage[utf8]{inputenc}

\usepackage{microtype}

\usepackage{inconsolata}

\usepackage{graphicx}

\title{StressRoBERTa: Cross-Condition Transfer Learning from Depression, Anxiety, and PTSD to Stress Detection}

\newcommand*{\affaddr}[1]{#1} 
\newcommand*{\affmark}[1][*]{\textsuperscript{#1}}
\newcommand*{\email}[1]{\texttt{#1}}

\author{%
\bf{Amal Alqahtani \affmark[1,2], Efsun Kayi \affmark[3], and Mona Diab\affmark[4]}\\
\affaddr{\affmark[1]The George Washington University, DC, USA}\\
\affaddr{\affmark[2]King Saud University, Riyadh, KSA}\\
\affaddr{\affmark[3]Johns Hopkins University Applied Physics Laboratory, Laurel, USA}\\
\affaddr{\affmark[4]Carnegie Mellon University, Pittsburgh, USA}\\
\email{amalqahtani@gwu.edu, efsun.kayi@jhuapl.edu, mdiab@andrew.cmu.edu}\\}

\begin{document}
\maketitle
\begin{abstract}
The prevalence of chronic stress represents a significant public health concern, with social media platforms like Twitter serving as important venues for individuals to share their experiences. This paper introduces StressRoBERTa, a cross-condition transfer learning approach for automatic detection of self-reported chronic stress in English tweets. The investigation examines whether continual training on clinically related conditions (depression, anxiety, PTSD), disorders with high comorbidity with chronic stress, improves stress detection compared to general language models and broad mental health models. RoBERTa is continually trained on the Stress-SMHD corpus (108M words from users with self-reported diagnoses of depression, anxiety, and PTSD) and fine-tuned on the SMM4H 2022 Task 8 dataset. StressRoBERTa achieves 82\% F1-score, outperforming the best shared task system (79\% F1) by 3 percentage points. The results demonstrate that focused cross-condition transfer from stress-related disorders (+1\% F1 over vanilla RoBERTa) provides stronger representations than general mental health training. Evaluation on Dreaddit (81\% F1) further demonstrates transfer from clinical mental health contexts to situational stress discussions.
\end{abstract}

\section{Introduction}
Chronic stress is a persistent sense of pressure that continues for an extended period \cite{vandenbos2007apa} and represents a significant public health concern. Understanding chronic stress is crucial given its serious negative effects on both physical and mental health. Numerous studies demonstrate that chronic stress can negatively affect the immune system \cite{khansari1990effects} and may lead to other mental illnesses such as depression or suicidality \cite{mcewen2006stress}. Detecting chronic stress is vital for preventing adverse health effects, implementing effective stress management techniques, addressing underlying causes, improving overall quality of life, and reducing costs related to healthcare and lost productivity.

Building on domain-adaptive training strategies, this paper investigates whether continual training on related mental health conditions improves the detection of a target condition. The approach is evaluated on the SMM4H 2022 Shared Task 8, which focuses on classifying self-reported chronic stress on Twitter \cite{weissenbacher-etal-2022-overview}. This task presents several challenges. First, the dataset is highly imbalanced (37\% positive, 63\% negative). Second, Twitter's 280-character limit requires models to extract stress signals from short text. Third, distinguishing genuine self-disclosure from mere mentions of stress requires understanding subtle linguistic cues. The shared task attracted multiple teams, with a median F1 score of 75\% and best performance of 79\% F1 \cite{huang-etal-2022-zydhjh4593}, indicating substantial task difficulty.

This work introduces a \textbf{cross-condition transfer learning} approach. Pretrained language models are continually trained on posts from users with depression, anxiety, and PTSD (conditions classified as stress-related disorders with high clinical comorbidity with chronic stress) and subsequently fine-tuned on stress detection. The central research question is whether continual training on a focused set of clinically related conditions (depression, anxiety, PTSD) improves stress detection compared to both general language models and broad mental health models.

The Stress-SMHD corpus \cite{cohan2018smhd} provides 108M words of training data from three stress-related conditions. This enables examination of whether continual training on a focused set of related conditions suffices for effective cross-condition transfer.

The contributions of this work are as follows:
\begin{enumerate}
  \item Competitive performance is achieved on SMM4H 2022 Shared Task 8 for stress detection on Twitter. StressRoBERTa attains 82\% F1, surpassing the best shared task system (79\% F1) by 3 percentage points and the median (75\% F1) by 7 percentage points.
  \item The results demonstrate that focused cross-condition continual training on stress-related disorders improves stress detection. StressRoBERTa outperforms vanilla RoBERTa-base by +1\% F1, while the general mental health model (MentalRoBERTa) shows no improvement, validating that condition selection matters.
  \item \emph{StressRoBERTa}\footnote{\url{https://huggingface.co/Amalq/stress-roberta-base}}, a continually trained language model achieving strong performance on Twitter (82\% F1, cross-platform transfer) and Dreaddit (81\% F1, demonstrating transfer from clinical to situational stress contexts), is presented.
  \item The cross-condition transfer approach is validated using clinical comorbidity evidence, linguistic overlap, and methodological comparisons, showing that a focused selection of stress-related disorders combined with self-reported diagnosis data provides a stronger signal than general mental health training.
\end{enumerate}

\section{Related Work}
\subsection{Continually Trained Large Language Models for Mental Health}
Continual training of pretrained language models for specific domains has advanced biomedical and healthcare natural language processing. \textbf{BioBERT} \cite{lee2020biobert} continually trains BERT on PubMed and PMC. \textbf{ClinicalBERT} \cite{alsentzer2019publicly} adapts BERT to clinical notes. \textbf{MentalBERT} and \textbf{MentalRoBERTa} \cite{ji-etal-2022-mentalbert} target mental health text from social media and continually train on mental health-related subreddits including r/depression, r/SuicideWatch, r/Anxiety, r/offmychest, r/bipolar, r/mentalillness, and r/mentalhealth.

\textbf{StressRoBERTa} uses focused cross-condition transfer and is continually trained specifically on stress-related disorders (r/depression, r/Anxiety, r/ptsd), selected for their high clinical comorbidity with chronic stress. This enables testing whether focused continual training on clinically related conditions outperforms general mental health continual training.

\subsection{SMM4H Shared Task 8: Stress Detection on Twitter}
The SMM4H 2022 Shared Task 8 \cite{weissenbacher-etal-2022-overview, Yang2022} focused on identifying self-reported chronic stress in English tweets. The task attracted multiple participating teams employing various approaches. \citet{huang-etal-2022-zydhjh4593} achieved the best performance (79\% F1) by fine-tuning BERT, RoBERTa, and BERTweet with pseudo-labeling in post-processing, finding that BERTweet outperforms BERT and RoBERTa. \citet{zhuang-zhang-2022-yet} explored multiple pretrained encoders including BioBERT \cite{lee2020biobert}, PubMedBERT \cite{gu2021domain}, DeBERTa \cite{he2020deberta}, and BERTweet \cite{nguyen2020bertweet} with varied loss functions, achieving 78\% F1. \citet{fu-etal-2022-casia-smm4h22} achieved 78\% F1, and \citet{kocaman-etal-2022-john} achieved 76\% F1. The median F1 across all submissions was 75\%.

To the authors' knowledge, no shared task systems employed cross-condition transfer learning in which a model is continually trained on related conditions (depression, anxiety, PTSD) and then fine-tuned on stress detection. Most systems used general pretrained language models (BERT, RoBERTa) or Twitter-specific models (BERTweet) without continual training on mental health data.

The use of cross-condition transfer for stress detection is supported by computational linguistics research. \citet{de2013predicting} demonstrated that stress-related linguistic markers appear consistently across depression, anxiety, and stress discussions. \citet{guntuku2017detecting} showed that models trained on depression and anxiety transfer effectively to stress detection, achieving F1 scores of 0.76 to 0.81.

\section{Method and Setup}

\begin{figure}[ht]
   \centering
      \includegraphics[width=.45\textwidth]{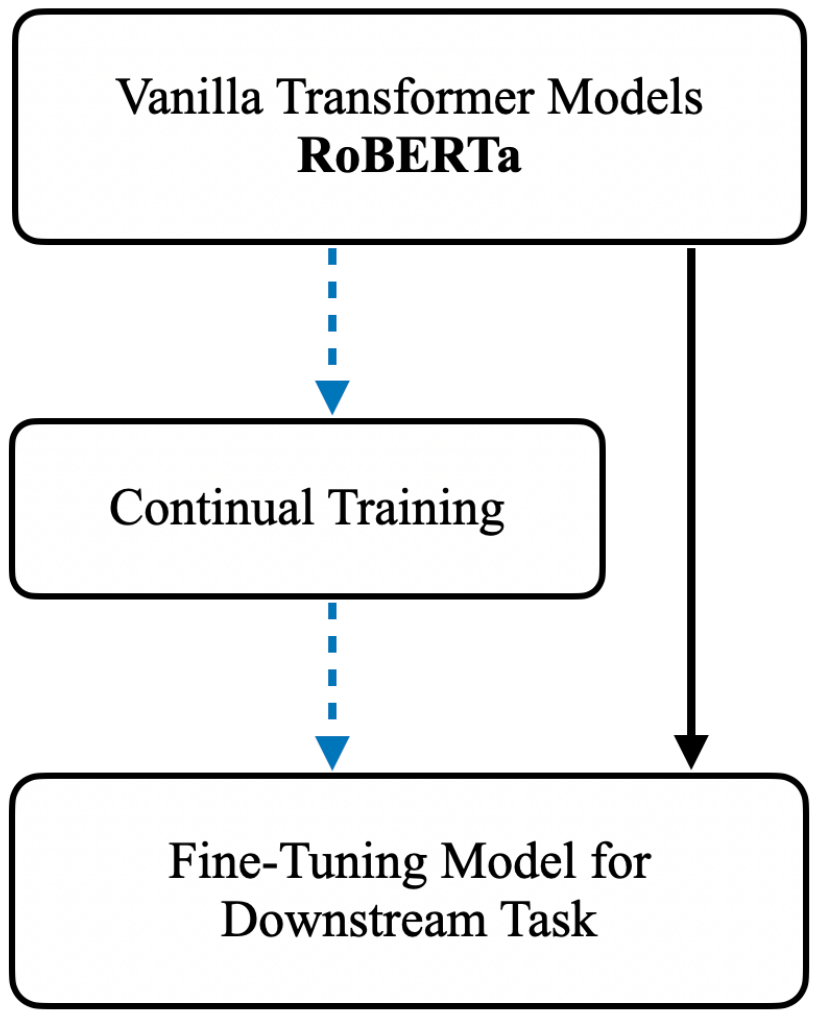}
 \caption{Overview of StressRoBERTa cross-condition transfer learning methodology}
 \label{fig:methodology1}
\end{figure}

\subsection{Language Model Continual Training}
The training follows standard RoBERTa protocols using the Transformers library \cite{wolf-etal-2020-transformers}. RoBERTa builds on the BERT architecture and is trained via masked language modeling \cite{liu2019roberta}. RoBERTa-base serves as the foundation model, with dynamic masking applied during training.

The training procedure follows domain-adaptive continual training \cite{gururangan2020don}. The model is initialized from the original RoBERTa checkpoint and continually trained on the Stress-SMHD corpus, which contains posts from users with depression, anxiety, and PTSD. This creates a cross-condition training scenario in which models learn representations from related mental health conditions, which are subsequently fine-tuned for stress detection. Figure~\ref{fig:methodology1} shows the overall process.

\begin{figure*}
    \centering
    \includegraphics[width=0.85\textwidth]{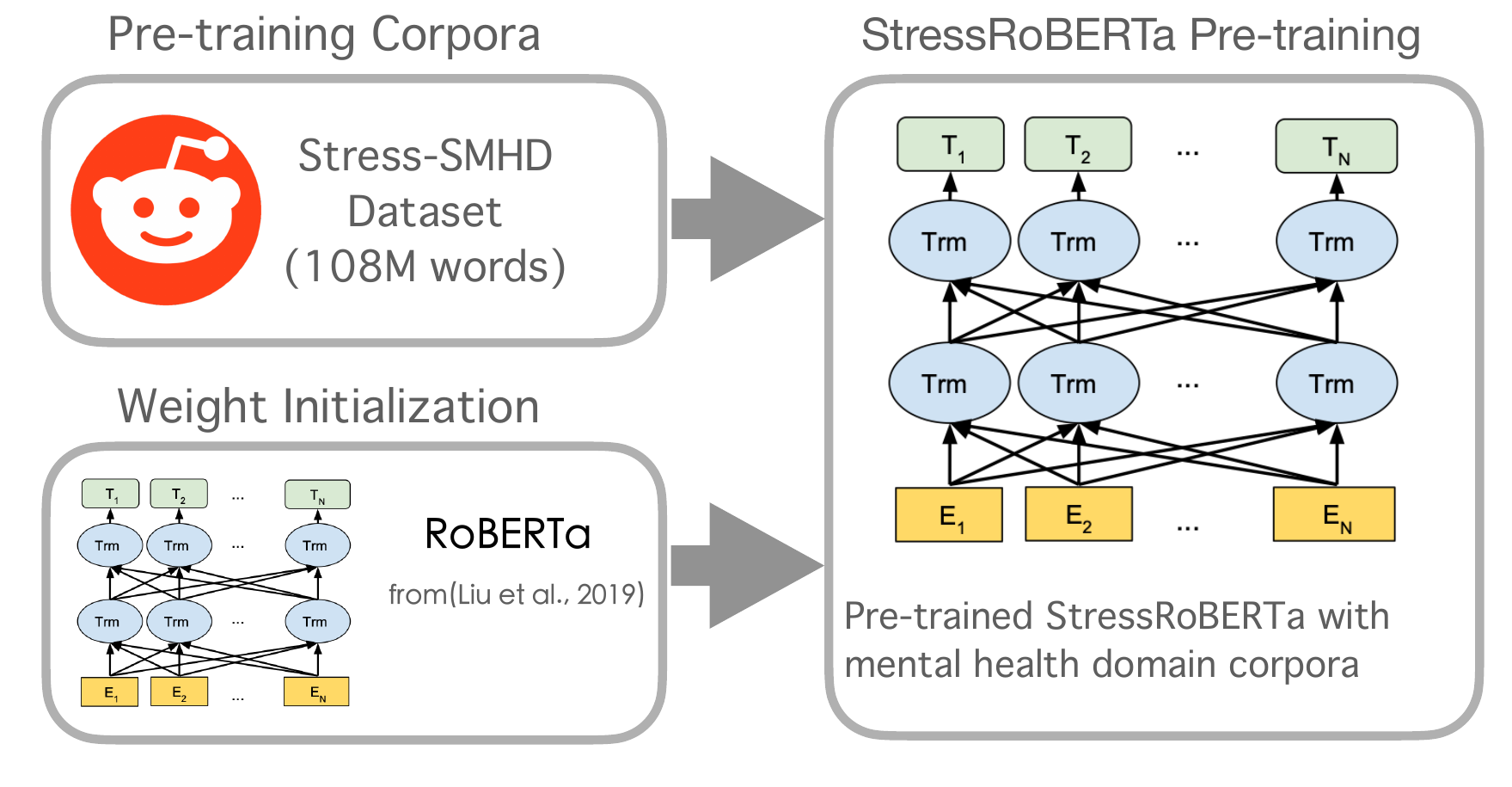}
    \caption{Detailed overview of StressRoBERTa cross-condition transfer learning pipeline}
    \label{fig:overview}
\end{figure*}

\subsection{Continual Training Corpus}

\subsubsection{Stress-SMHD}
Stress-SMHD is a subset of the Self-Reported Mental Health Diagnoses dataset \cite{cohan2018smhd} containing Reddit posts from users with diagnoses of anxiety, PTSD, or depression. This ensures the continual training data reflects language from individuals with these conditions. The corpus does not contain explicit stress labels but consists of posts discussing experiences with depression, anxiety, and PTSD. This enables evaluation of whether representations learned from these related conditions transfer effectively to stress detection. Tables~\ref{tab:continual_training_corpus_stress} and~\ref{tab:stress_corpus_statistics} summarize the corpus composition.

\begin{table}[ht]
\centering
\begin{tabular}{lcc}
\hline
\textbf{Corpus} & \textbf{Num of Words} & \textbf{Domain}\\
\hline
English Wikipedia & 2.5B & General\\
BooksCorpus & 0.8B & General\\
CC-News & 10B & General\\
\hline
\textbf{Stress-SMHD} & \textbf{108M} & \textbf{Psych.}\\
\hline
\end{tabular}
\caption{Text Corpora Used for StressRoBERTa Continual Training}
\label{tab:continual_training_corpus_stress}
\end{table}

\begin{table}[ht]
\centering
\begin{tabular}{lcc}
\hline
\textbf{Condition} & \textbf{Posts} & \textbf{Tokens} \\
\hline
Depression & 1,272k & 57.4M (53\%)\\
Anxiety & 795k & 36.9M (34\%)\\
PTSD & 258k & 13.7M (13\%)\\\hline
\textbf{Total} & \textbf{2,325k} & \textbf{108M (100\%)}\\
\hline
\end{tabular}
\caption{Composition of Stress-SMHD Corpus by Condition}
\label{tab:stress_corpus_statistics}
\end{table}

\subsection{StressRoBERTa Model}
StressRoBERTa is initialized with RoBERTa-base weights \cite{liu2019roberta} pretrained on English Wikipedia, BooksCorpus, and CC-News. Continual training is performed on Stress-SMHD using the Transformers library \cite{wolf-etal-2020-transformers}, adapting the model to learn representations from depression, anxiety, and PTSD discussions.

\subsection{Continual Training Configuration}
StressRoBERTa is continually trained on Stress-SMHD using the Huggingface Transformers library. Table~\ref{tab:continual_training_config} presents the training configuration. Masked language modeling with 15\% dynamic masking is applied following \cite{liu2019roberta}. The model achieves a final perplexity of 5.22 on the Stress-SMHD corpus.

\begin{table}[ht]
\centering
\begin{tabular}{lc}
\hline
\textbf{Parameter} & \textbf{Value}  \\
\hline
Base model & RoBERTa-base\\
Learning rate & $2 \times 10^{-5}$\\
Batch size & 16\\
Number of epochs & 5\\
Weight decay & 0.01\\
Max sequence length & 512\\
Final perplexity & 5.22\\
\hline
\end{tabular}
\caption{Continual Training Hyperparameters}
\label{tab:continual_training_config}
\end{table}
\section{Evaluation}

\subsection{Datasets}

\subsubsection{SMM4H 2022 Shared Task 8: Twitter Stress Detection}
Evaluation is conducted on the corpus from SMM4H 2022 Shared Task 8 \cite{weissenbacher-etal-2022-overview}, which focuses on identifying self-disclosures of chronic stress in English tweets. This binary classification task distinguishes tweets in which users explicitly report experiencing chronic stress from those that mention stress without self-disclosure. The task is challenging due to class imbalance (37\% positive, 63\% negative), short text length (280-character Twitter limit), and the need to distinguish genuine self-disclosure from other stress mentions. Table~\ref{tab:smm4h_distribution} reports the label distribution.

\begin{table}[ht]
\centering
\begin{tabular}{lllc}
\hline
\textbf{Split} & \textbf{P} & \textbf{N} & \textbf{\#} \\
\hline
Train & 1,092 (37\%) & 1,844 (63\%) & 2,936 \\
Valid & 156 (37\%) & 264 (63\%) & 420 \\
Test & NA & NA & 839 \\\hline
\end{tabular}
\caption{Distribution of SMM4H 2022 Task 8 Twitter Stress Detection Dataset}
\label{tab:smm4h_distribution}
\end{table}

\subsubsection{Dreaddit: Reddit Stress Detection}
To evaluate transfer from clinical mental health contexts to situational stress discussions, testing is also conducted on Dreaddit \cite{turcan2019dreaddit}, a Reddit dataset for stress analysis containing multi-domain posts labeled for binary stress classification. Posts are sourced from ten subreddits representing diverse stress contexts. The training split has 2,838 posts (52\% stress). The test split has 715 posts.

\subsection{Experimental Setup}
RoBERTa-base \cite{liu2019roberta} is fine-tuned as the general baseline and StressRoBERTa as the cross-condition model. Comparisons are also made against domain-specific baselines, including ClinicalBERT \cite{alsentzer2019publicly}, BioBERT \cite{lee2020biobert}, MentalBERT, and MentalRoBERTa \cite{ji-etal-2022-mentalbert}. All models are base-sized for fair comparison and share the configuration in Table~\ref{tab:finetuning_config}.

\begin{table}[ht]
\centering
\begin{tabular}{lc}
\hline
\textbf{Parameter} & \textbf{Value}  \\
\hline
Learning rate & $2e^{-5}$\\
Train batch size & 16\\
Validation batch size & 16\\
Number of train epochs & 6 \\
Weight decay & 0.01\\
Early stopping patience & 3 epochs\\\hline
\end{tabular}
\caption{Fine-Tuning Hyperparameters}
\label{tab:finetuning_config}
\end{table}

\section{Results}
\subsection{Performance on SMM4H 2022 Shared Task 8}
Table~\ref{tab:smm4h_results} reports F1 and recall for the positive (stress) class, comparing the proposed approach with shared-task participants and baseline models.

StressRoBERTa achieves 82\% F1 on the Twitter stress detection task, outperforming all shared task participants and baseline models. This represents a 3 percentage point improvement over the best shared task system \cite{huang-etal-2022-zydhjh4593} (79\% F1) and a 7 percentage point gain over the shared task median (75\% F1). The results demonstrate that cross-condition transfer learning from stress-related disorders provides effective representations for stress detection in short social media text. Future work should include statistical significance testing across multiple random seeds to validate the robustness of these improvements.

\subsubsection{Comparison to Shared Task Systems}
The proposed approach differs from shared-task systems in several key ways. First, shared task systems primarily used general pretrained language models (BERT, RoBERTa) or Twitter-specific models (BERTweet) without domain-adaptive continual training on mental health data. The best performing system \cite{huang-etal-2022-zydhjh4593} used ensemble methods and pseudo-labeling post-processing rather than domain-specific continual training. Second, no shared task systems employed cross-condition transfer learning from related mental health conditions.

StressRoBERTa's superior performance suggests that domain-adaptive continual training on stress-related disorders captures linguistic patterns relevant to stress detection that general pretrained language models miss. The focused selection of depression, anxiety, and PTSD (conditions with high clinical comorbidity with stress) provides a stronger signal than either general language models or Twitter-specific models trained on diverse content.

\subsubsection{Comparison to Domain-Specific Baselines}

StressRoBERTa also outperforms domain-specific baseline models. Clinical domain models (ClinicalBERT 75\% F1, BioBERT 80\% F1) perform poorly, suggesting that clinical note language does not transfer well to social media stress detection. The comparison to general mental health models is more relevant. MentalBERT achieves 81\% F1 and MentalRoBERTa achieves 81\% F1, identical to vanilla RoBERTa-base (81\% F1). This reveals a critical insight. General mental health continual training provides no improvement for stress detection, while focused cross-condition continual training on stress-related disorders improves performance by +1\% F1.

Both StressRoBERTa and MentalRoBERTa use the RoBERTa-base architecture with identical fine-tuning procedures but differ in continual training corpus selection. MentalRoBERTa is continually trained on general mental health subreddits (r/depression, r/SuicideWatch, r/Anxiety, r/offmychest, r/bipolar, r/mentalillness, r/mentalhealth). StressRoBERTa is continually trained on stress-related disorders (r/depression, r/Anxiety, r/ptsd). MentalRoBERTa's lack of improvement over vanilla RoBERTa-base suggests that including diverse mental health content reduces the effectiveness of transfer learning for stress detection. StressRoBERTa's focused selection provides more substantial clinical and linguistic overlap with stress, enabling effective cross-condition transfer.

\begin{table*}[htbp]
\centering
\begin{tabular}{lcc}
\hline
\textbf{Models} & \textbf{Recall} & \textbf{F1}\\ \hline
\multicolumn{3}{l}{\textit{SMM4H 2022 Shared Task 8 Participants}}\\
Huang et al. (Best system) \cite{huang-etal-2022-zydhjh4593} & 85\% & 79\%\\
Zhuang and Zhang \cite{zhuang-zhang-2022-yet} & 76\% & 78\%\\
Fu et al. \cite{fu-etal-2022-casia-smm4h22} & 82\% & 78\%\\
Kocaman et al. \cite{kocaman-etal-2022-john} & \textbf{87\%} & 76\%\\
Median of all submissions & 76\% & 75\%\\\hline
\multicolumn{3}{l}{\textit{General language models}}\\
BERT-base-uncased & 75\% & 80\%\\
RoBERTa-base & 83\% & 81\%\\\hline
\multicolumn{3}{l}{\textit{Domain-Specific language models}}\\
ClinicalBERT \cite{alsentzer2019publicly} & 64\% & 75\%\\
BioBERT-base \cite{lee2020biobert} & 77\% & 80\%\\
MentalBERT \cite{ji-etal-2022-mentalbert} & 77\% & 81\%\\
MentalRoBERTa \cite{ji-etal-2022-mentalbert} & 85\% & 81\%\\\hline
\multicolumn{3}{l}{\textit{Proposed Approach (Cross-Condition Transfer)}}\\
\textbf{StressRoBERTa} & 84\% & \textbf{82\%}\\
\hline
\end{tabular}
\caption{Performance Comparison on SMM4H 2022 Task 8 Twitter Stress Detection. StressRoBERTa achieves 82\% F1, outperforming the best shared task system (79\% F1) and all baseline models.}
\label{tab:smm4h_results}
\end{table*}

\subsection{Cross-Platform and Cross-Context Generalization}
To assess generalizability, two complementary evaluations are conducted. First, cross-platform transfer from Reddit to Twitter is tested using the SMM4H Task 8 dataset. Second, cross-context transfer from clinical/diagnostic subreddits (r/depression, r/Anxiety, r/ptsd) to situational stress discussions is tested using Dreaddit, which comprises posts from diverse topical communities. Table~\ref{tab:platform_comparison} shows performance on both benchmarks.

\begin{table}[ht]
\centering
\begin{tabular}{lcc}
\hline
\textbf{Model} & \textbf{Twitter} & \textbf{Dreaddit}\\
\hline
BERT-base & 0.80 & 0.79\\
RoBERTa-base & 0.81 & 0.81\\
MentalRoBERTa & 0.81 & 0.81\\
\textbf{StressRoBERTa} & \textbf{0.82} & \textbf{0.81}\\
\hline
\end{tabular}
\caption{Performance (F1 scores) on Twitter and Reddit. Twitter evaluates cross-platform transfer while Dreaddit evaluates cross-context transfer.}
\label{tab:platform_comparison}
\end{table}

StressRoBERTa achieves 81\% F1 on Dreaddit, demonstrating that representations learned from clinical mental health discussions (depression, anxiety, PTSD) transfer effectively to situational stress contexts. The consistent performance across Twitter (82\% F1, cross-platform) and Dreaddit (81\% F1, cross-context) validates that cross-condition continual training captures transferable stress-related patterns rather than platform-specific or context-specific artifacts.

\subsection{Analysis and Discussion}

\subsubsection{Why Does Cross-Condition Transfer Improve Performance?}

Three factors explain StressRoBERTa's superior performance on the stress detection task.

\paragraph{Clinical Comorbidity} High clinical comorbidity rates support the validity of cross-condition transfer. Studies show 60 to 80\% of individuals with depression report chronic stress \cite{mazure1998life, tennant2002life}, and 50 to 70\% of anxiety patients meet criteria for chronic stress \cite{kessler2013effects}. Chronic stress increases depression risk 2.5 to 3.0 fold \cite{kendler1999causal} and anxiety risk 2.0 to 2.8 fold \cite{kessler2013effects}. This overlap means language from individuals discussing depression, anxiety, or PTSD inherently contains stress-related content, making these conditions appropriate source domains for stress detection.

\paragraph{Linguistic Overlap} Computational studies validate that stress markers appear consistently across these conditions. \citet{de2013predicting} found stress-related terms in depression (42\%), anxiety (51\%), and PTSD (38\%) discussions with similar frequencies. \citet{guntuku2017detecting} demonstrated successful cross-condition transfer with an accuracy of 0.78 to 0.82, comparable to the present results, showing that stress markers (first-person pronouns, negative emotion words, present-tense verbs) appear consistently across source and target conditions.

\paragraph{Focused Condition Selection} Two methodological factors strengthen cross-condition transfer. Both StressRoBERTa and MentalRoBERTa include depression and anxiety subreddits. However, StressRoBERTa adds r/ptsd, a stress-related disorder in which 50 to 70\% of patients meet criteria for chronic stress, while MentalRoBERTa adds broader mental health content (r/mentalillness, r/mentalhealth, r/offmychest, r/bipolar, r/SuicideWatch). Additionally, Stress-SMHD comprises posts from users who explicitly self-reported diagnoses, ensuring that the language reflects individuals who identify as having these conditions. MentalRoBERTa was trained on all posts from mental health subreddits, including general discussions and advice-seeking, without requiring self-reported diagnoses. This suggests that focused cross-condition transfer benefits from both selecting clinically related conditions and using self-reported diagnosis data.

Table~\ref{tab:corpus_comparison} compares the training corpora for StressRoBERTa and MentalRoBERTa, highlighting the key differences in corpus selection strategy.

\begin{table*}[ht]
\centering
\begin{tabular}{lp{6cm}cc}
\hline
\textbf{Model} & \textbf{Training Conditions} & \textbf{Self-Reported?} & \textbf{Tokens} \\
\hline
MentalRoBERTa & depression, anxiety, suicide, bipolar, general mental health & No & ~150M\\
\textbf{StressRoBERTa} & \textbf{depression, anxiety, PTSD} & \textbf{Yes} & \textbf{108M}\\
\hline
\end{tabular}
\caption{Comparison of continual training corpora. StressRoBERTa uses a focused selection of stress-related disorders with self-reported diagnoses, while MentalRoBERTa includes broader mental health content without diagnosis requirements.}
\label{tab:corpus_comparison}
\end{table*}

\subsubsection{Comparison to Prior Cross-Condition Transfer Work}

The results (82\% F1) are comparable to prior cross-condition transfer studies. \citet{guntuku2017detecting} achieved 0.76 to 0.81 F1 transferring from depression and anxiety to stress detection, validating the feasibility of cross-condition approaches. This work extends these findings by demonstrating that focused condition selection based on clinical relationships outperforms general mental health continual training and achieves competitive performance on a challenging shared task benchmark.

\section{Conclusion}

This work demonstrates that cross-condition transfer learning from stress-related disorders improves stress detection on social media. StressRoBERTa achieves 82\% F1 on SMM4H 2022 Shared Task 8, outperforming the best shared task system (79\% F1) by 3 percentage points and the median (75\% F1) by 7 percentage points.

The primary contribution lies in demonstrating that focused cross-condition continual training on clinically related conditions (depression, anxiety, PTSD) outperforms both general pretrained language models and general mental health models. While vanilla RoBERTa-base achieves 81\% F1, the general mental health model (MentalRoBERTa) shows no improvement (81\% F1), and focused cross-condition continual training improves performance to 82\% F1. This validates that condition selection based on clinical relationships matters for cross-condition transfer.

Clinical comorbidity and linguistic overlap between source conditions and the target condition enable effective knowledge transfer, supported by empirical results across Twitter (82\% F1, cross-platform) and Dreaddit (81\% F1, cross-context). These findings have important implications for mental health NLP. Domain-adaptive continual training on carefully selected related conditions provides stronger representations than either general language models or broad mental health continual training.

\section*{Limitations}
StressRoBERTa demonstrates strong performance for English stress detection on social media but has several limitations. First, the model is trained and evaluated exclusively on English text from social media platforms (Reddit and Twitter). Performance on other languages, different domains (e.g., clinical notes, formal writing), and longer documents remains unexplored. Second, the Stress-SMHD corpus comprises only posts from three subreddits (r/depression, r/Anxiety, r/ptsd) with self-reported diagnoses, which may not capture the full linguistic diversity of stress-related discourse across different populations and contexts. Third, no systematic evaluation was conducted of alternative condition combinations for cross-condition transfer. Fourth, training requires substantial computational resources, though the pretrained model will be released to improve accessibility. Finally, as with any automated mental health detection system, StressRoBERTa should only be deployed with appropriate ethical safeguards and clinical oversight to prevent potential misuse.

\section*{Acknowledgments}
During the preparation of this work, AI assistants were used to correct grammar and improve the clarity of writing.

\section*{Ethics Statement}
This work involves automated detection of mental health conditions from social media data, which raises several ethical considerations.

\paragraph{Privacy and Consent} The continual training uses the Stress-SMHD dataset derived from Reddit, where users posted publicly. However, users may not have anticipated their posts being used for mental health research. No new data was collected and no attempts were made to identify individuals.

\paragraph{Clinical Validation} The model achieves 82\% F1, meaning approximately 18\% of cases are misclassified. False negatives could result in missed opportunities for intervention, while false positives could cause unnecessary concern. Any deployment must involve human clinical judgment and should not replace professional mental health assessment.

\paragraph{Responsible Release} StressRoBERTa is released to advance mental health research, but users are urged to consider these ethical implications and implement appropriate safeguards in any applications.

\bibliography{custom}

\end{document}